\documentclass[lettersize,journal]{IEEEtran}
\usepackage{amsmath,amsfonts}
\usepackage{algorithmic}
\usepackage{algorithm}
\usepackage{array}
\usepackage[caption=false,font=normalsize,labelfont=sf,textfont=sf]{subfig}
\usepackage{textcomp}
\usepackage{stfloats}
\usepackage{url}
\usepackage{verbatim}
\usepackage{graphicx}
\usepackage{cite}
\usepackage{cite}
\usepackage{booktabs}
\usepackage{float}
\usepackage{orcidlink}
\usepackage{mdframed}
\usepackage{xcolor}
\newmdenv[backgroundcolor=gray!10, linecolor=black, linewidth=0.5pt, roundcorner=4pt, innerleftmargin=2pt, innerrightmargin=2pt, innertopmargin=5pt, innerbottommargin=5pt]{exampleprompt}
\begin{document}

\title{From Shots to Stories: LLM-Assisted Video Editing with Unified Language Representations}

\author{
  Yuzhi Li\orcidlink{0009-0009-7050-7775}, Haojun Xu\orcidlink{0000-0001-8312-6634}, Feng Tian\orcidlink{0009-0003-9904-6729}

        % <-this % stops a space
\thanks{\textit{Corresponding Author: Feng Tian}}
\thanks{Yuzhi Li, Haojun Xu and Feng Tian are with Shanghai University, Shanghai, China, 200072, (e-mails: shadowmcv@shu.edu.cn; sacross\_k@shu.edu.cn; ouman@shu.edu.cn)}
\thanks{This work will be submitted to the IEEE or other Press for possible publication. Copyright may be transferred without notice, after which this version may no longer be accessible.}
% \thanks{Manuscript received April 19, 2021; revised August 16, 2021.}
}
\markboth{Journal of \LaTeX\ Class Files,~Vol.~14, No.~8, August~2021}%
{Shell \MakeLowercase{\textit{et al.}}: A Sample Article Using IEEEtran.cls for IEEE Journals}

\IEEEpubid{}
% Remember, if you use this you must call \IEEEpubidadjcol in the second
% column for its text to clear the IEEEpubid mark.

\maketitle

\begin{abstract}
Large Language Models (LLMs) and Vision-Language Models (VLMs) have demonstrated remarkable reasoning and generalization capabilities in video understanding; however, their application in video editing remains largely underexplored. This paper presents the first systematic study of LLMs in the context of video editing. To bridge the gap between visual information and language-based reasoning, we introduce L-Storyboard, an intermediate representation that transforms discrete video shots into structured language descriptions suitable for LLM processing. We categorize video editing tasks into Convergent Tasks and Divergent Tasks, focusing on three core tasks: Shot Attributes Classification, Next Shot Selection, and Shot Sequence Ordering. To address the inherent instability of divergent task outputs, we propose the StoryFlow strategy, which converts the divergent multi-path reasoning process into a convergent selection mechanism, effectively enhancing task accuracy and logical coherence. Experimental results demonstrate that L-Storyboard facilitates a more robust mapping between visual information and language descriptions, significantly improving the interpretability and privacy protection of video editing tasks. Furthermore, StoryFlow enhances the logical consistency and output stability in Shot Sequence Ordering, underscoring the substantial potential of LLMs in intelligent video editing.
\end{abstract}


\section{Introduction}

Large Language Models (LLMs) \cite{cite9,cite10} and Vision-Language Models (VLMs) \cite{cite32,cite33} have demonstrated remarkable reasoning and generalization capabilities in video understanding. However, their potential applications in video editing tasks remain largely underexplored. With the rapid proliferation of social media and online streaming platforms, the production and consumption of video content have grown exponentially, creating an urgent demand for efficient, intelligent, and user-friendly video editing technologies \cite{cite4}. 

Traditional deep learning frameworks have long dominated the video editing domain, primarily treating various editing tasks as independent computer vision problems \cite{cite5,cite6,cite7}. This isolated approach often lacks explicit connections to high-level semantic information or editorial intent. Moreover, compared to fundamental tasks like video classification, these editing tasks exhibit significant shortcomings in interpretability \cite{cite8}, making it challenging to intuitively present the logic and reasoning behind editorial decisions.

\begin{figure*}[ht]
 \centering 
 \includegraphics[width=2\columnwidth]{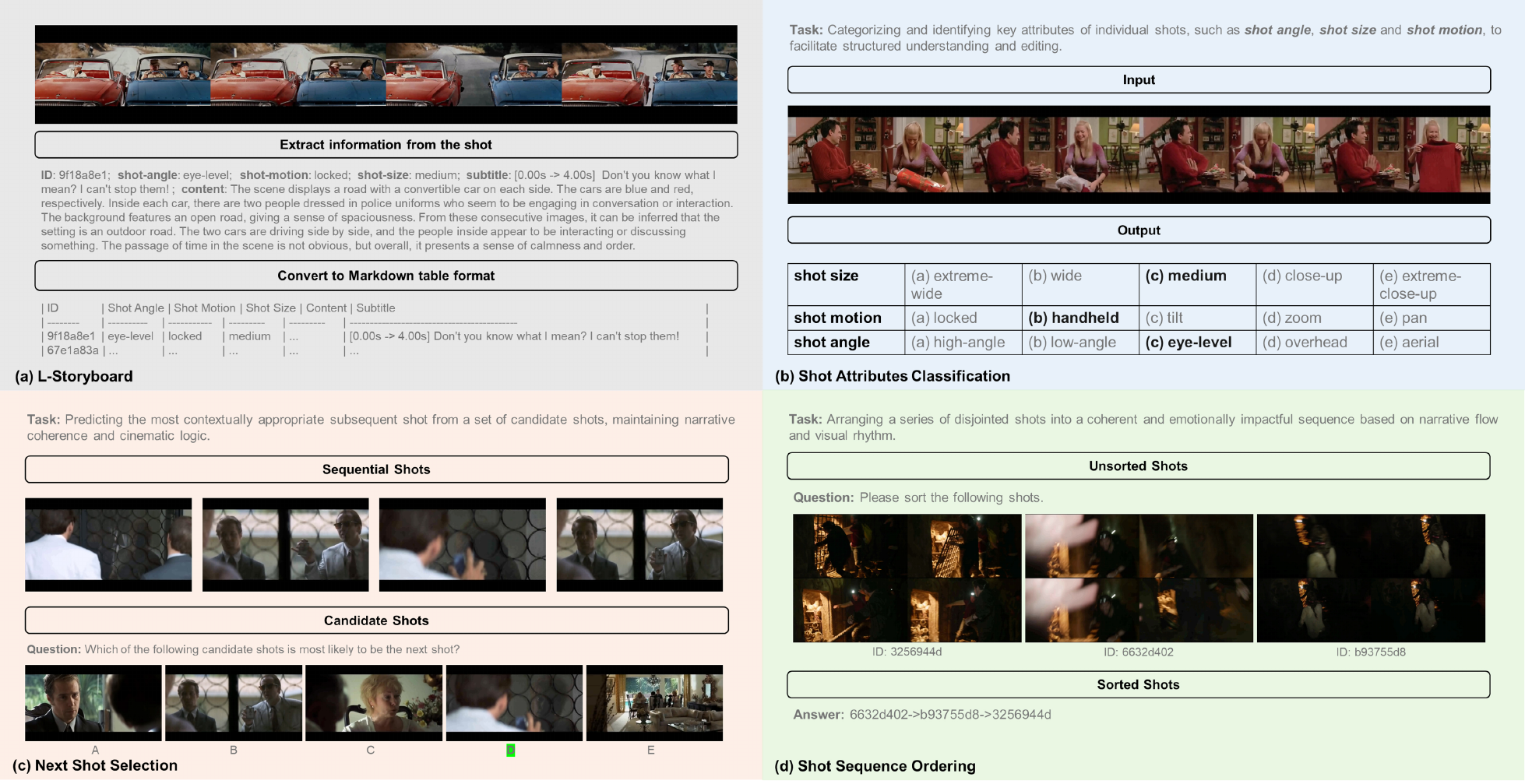}
 \caption{Overview of our proposed L-Storyboard and AI-assisted video editing tasks. (a) L-Storyboard (b) Shot Attributes Classification (c) Next Shot Selection (d) Shot Sequence Ordering}
 \label{fig:1}
\end{figure*}

To address these limitations, this study explores how LLMs can systematically enhance video editing performance, introducing a novel framework termed LLM-Assisted Video Editing. As illustrated in Figure \ref{fig:1}, inspired by traditional film storyboarding techniques \cite{cite11}, we designed an intermediate representation called L-Storyboard. This representation transforms discrete video shots into structured language descriptions that LLMs can effectively process, enabling the mapping of visual information into the language domain. Building on this representation, we further investigate three foundational AI-assisted video editing tasks: Shot Attributes Classification, Next Shot Selection, and Shot Sequence Ordering \cite{cite1}. These three tasks collectively encompass key aspects of video editing and form the foundation for practical applications.

To thoroughly explore the performance of LLMs across different tasks, we categorize video editing tasks into two primary types: Convergent Tasks and Divergent Tasks. Convergent Tasks require reasoning that "converges" from multiple possibilities to an almost singular optimal solution, whereas Divergent Tasks involve "diverging" into multiple reasonable possibilities from limited inputs, resulting in a broader solution space and greater diversity.

Specifically, Shot Attributes Classification \cite{cite5} focuses on extracting structured shot language from individual scenes, while Next Shot Selection \cite{cite6} identifies the most suitable subsequent shot from a set of candidates. These two tasks typically have clear optimal solutions, classifying them as Convergent Tasks. In contrast, Shot Sequence Ordering \cite{cite7} involves arranging multiple shots into coherent segments based on narrative logic and emotional rhythm. Given its requirement to generate various plausible sequences, it is considered a typical Divergent Task. Existing research indicates that deep learning methods generally excel in Convergent Tasks compared to Divergent Tasks, with the latter exhibiting significantly lower accuracy rates \cite{cite1}.

Regarding interpretability, Shot Attributes Classification can be viewed as a specialized form of traditional video classification, capable of leveraging methods like Grad-CAM \cite{cite2} or explicit visual cues \cite{cite3} for intuitive explanations. However, Next Shot Selection and Shot Sequence Ordering lack equivalent interpretive mechanisms. The step-by-step reasoning capabilities of LLMs \cite{cite12} offer a natural, human-readable explanation process, facilitating better interpretation of complex reasoning pathways. 

To address the challenges associated with the Divergent Task of Shot Sequence Ordering, we propose a method called StoryFlow. This approach first utilizes LLMs with varying degrees of creativity (temperature values) to generate multiple candidate story sequences. Subsequently, a preset evaluation criterion is applied to select the sequence that best matches narrative logic, effectively transforming the latter half of the Divergent Task into a Convergent selection problem.

Experimental results demonstrate that the L-Storyboard representation significantly enhances the performance of LLMs (including VLMs) in video editing tasks. For Shot Attributes Classification and Next Shot Selection, LLMs outperform traditional deep learning methods under single-turn dialogue conditions. However, in the more complex Shot Sequence Ordering task, directly applying LLMs or fine-tuned versions resulted in decreased accuracy, revealing current limitations in handling multi-path reasoning and complex sequence construction. To address these challenges, StoryFlow is introduced as an optimization module, effectively improving performance in shot sequencing. Compared to traditional deep learning methods, this mechanism not only enhances sorting accuracy but also tackles the long-standing interpretability challenges in video editing tasks.

The main contributions of this paper are summarized as follows:

\begin{itemize}
    \item[1.] We propose L-Storyboard, a method that maps shot-level visual information into a language modality, providing LLMs with a unified and efficient input representation for video editing tasks.
    \item[2.] From the perspective of information flow, we systematically categorize video editing tasks into Convergent and Divergent types, and explore these categories through three core foundational tasks.
    \item[3.] To address optimization challenges in Divergent Tasks, we design and validate the StoryFlow strategy, significantly improving the performance and logical consistency of Shot Sequence Ordering.
    \item[4.] Through extensive experimentation, we evaluate the performance of LLM/VLM models in key video editing tasks, offering new perspectives and reference benchmarks for future research.
   
\end{itemize}

\section{Related Work}

\subsection{AI-Assisted Video Editing}
The development of AI-assisted video editing technologies has primarily progressed along two technical pathways: visual-level understanding and timeline editing. In terms of visual-level understanding, researchers have focused on the in-depth analysis of shot content and its corresponding shot language, encompassing tasks such as shot language comprehension \cite{cite5}\cite{cite15}, character recognition \cite{cite13}, and character relationship construction \cite{cite14}. In recent years, the semantic understanding of shots has emerged as a central topic within this domain. This task is often regarded as a form of video understanding and can be integrated with mainstream interpretability techniques for result visualization \cite{cite4}\cite{cite16}. For instance, Rao et al. proposed SGNet \cite{cite5}, which employs a subject-background separation method to extract hierarchical shot language information from scenes, enabling finer-grained comprehension of shot content. Similarly, SCTSNet \cite{cite17}, introduced by Jiang et al., combines shot language understanding with shot boundary detection, enhancing both shot interpretation and the accuracy of transitions between scenes. Moreover, related studies have identified potential shared feature spaces among different shot languages \cite{cite2}, facilitating unified modeling and analysis of complex shot expressions.

In the realm of timeline editing, research has primarily concentrated on the organization and scheduling of shots along the video timeline, encompassing tasks such as video summarization \cite{cite18}, trailer generation \cite{cite20}, and shot sequence ordering \cite{cite7}. To improve both efficiency and quality, numerous visual editing tools have been developed, including Write A Video \cite{cite21} and QuickCut \cite{cite22}. These tools are designed to quickly extract video content and efficiently reorganize it. Additionally, GraphTrailer \cite{cite19}, introduced by Papalampidi et al., utilizes a graph-based structure to represent shots as nodes and their relationships as edges. This graph-based representation, combined with traversal algorithms, selects the most appropriate shots for trailer generation, thereby significantly improving the coherence and logical flow of video summarization and editing.

Although these task-specific technologies have achieved considerable advancements in their respective areas, they still face significant challenges. These include the lack of collaborative optimization across tasks, which hinders the formation of a unified editing logic, and the inability to explicitly express user editing intentions, which limits the fluidity of human-computer interaction.

To address these challenges, this paper introduces a generalized shot representation method based on LLMs, termed L-Storyboard. By converting discrete shots into structured language descriptions that LLMs can process, this approach maps visual information into the language modality, enabling video editing tasks to be reasoned about and executed within a unified language space. Compared to traditional approaches that require separate models for each task, this unified representation fully leverages the reasoning and generative capabilities of pre-trained LLMs, reducing both information loss and redundant computation across multi-task scenarios. Furthermore, we select three fundamental video editing tasks—Shot Attributes Classification, Next Shot Selection, and Shot Sequence Ordering—to validate the application potential of LLMs in AI-assisted video editing.

\subsection{LLM-Powered Video Understanding}

LLMs, particularly VLMs, are reshaping the technological paradigm of video understanding. Traditional video understanding tasks typically rely on deep learning models to perform isolated analyses of visual features. In contrast, VLMs facilitate the deep integration of visual and language modalities, enabling the comprehension and reasoning of complex video semantics. For example, GPT-4o \cite{cite23} effectively bridges the gap between vision and language, achieving high-precision video content parsing even under zero-shot learning conditions \cite{cite24}\cite{cite25}. This capability allows VLMs not only to detect and describe explicit objects, scenes, and actions within video shots but also to perform high-level semantic reasoning about complex narrative logic, character relationships, and emotional expressions.

In specific applications, Ranasinghe et al. introduced the Multimodal Video Understanding (MVU) \cite{cite26} framework, which integrates global object information, spatial positioning, and cross-shot temporal information. By leveraging the language reasoning capabilities of LLMs, MVU achieves efficient understanding and content extraction from long-duration videos. Similarly, Min et al. proposed MoReVQA \cite{cite27}, which incorporates a multi-stage, modular reasoning framework into video question-answering tasks, significantly enhancing performance in complex video Q\&A scenarios while improving the interpretability of the reasoning process.

To address the challenge of redundant information in long videos, some VLM frameworks have introduced memory-enhanced or hierarchical understanding strategies to improve the efficiency and accuracy of long-sequence video reasoning. For example, Video Agent \cite{cite28}, proposed by Fan et al., segments videos into short time intervals, independently generating semantic descriptions for each segment while annotating timestamps and event information. It also introduces a Re-ID mechanism to eliminate redundant objects, optimizing cross-timeline information matching and retrieval. Meanwhile, VideoTree \cite{cite29}, introduced by Wang et al., employs visual clustering techniques to partition long videos into a tree structure based on relevance. During inference, it reads information top-down, layer by layer, to construct structured textual descriptions.
In the field of video editing, LLM-based automated editing tools are becoming research hotspots. For instance, LAVE \cite{cite30}, proposed by Wang et al., allows for the automatic adjustment of shot positions and order on the timeline through natural language commands, enabling rapid, semantic-based reorganization and editing. ChunkyEdit \cite{cite31}, developed by Leake et al., targets the editing of long interview videos, automatically generating rough-cut versions in Adobe Premiere while supporting secondary edits based on textual descriptions. These studies demonstrate how the reasoning and generative capabilities of VLMs are transforming video editing interactions, significantly enhancing the operational efficiency for non-professional users.

Despite these advancements, existing VLMs still face notable limitations in practical applications. First, their results tend to be highly variable and lack stability when handling divergent tasks such as shot sequencing. Second, with regard to privacy protection, many VLM-based editing tools require cloud-based processing, which raises potential risks of privacy leakage, particularly when dealing with personal visual information.

To overcome these challenges, this paper introduces the StoryFlow method, which transforms divergent tasks into more deterministic convergent problems, significantly improving the accuracy and coherence of sequencing tasks. Furthermore, the L-Storyboard representation serves as an anonymized visual description method, preventing the exposure of character identities during processing. For instance, using expressions like "the person in white clothes" instead of explicit character images effectively reduces privacy risks. Additionally, the models employed in the experimental design are optimized to run on consumer-grade personal computers, including quantized versions, thereby lowering hardware requirements and enhancing the feasibility of practical deployment.

\section{L-Storyboard: Bridging Video Editing Tasks with LLMs}

\subsection{Design Motivation}
In traditional filmmaking, the Storyboard serves as a crucial tool for directors and editors to intuitively visualize shots, scenes, and narrative sequences \cite{cite11}. Through graphical representation, it concretizes complex narrative logic and shot arrangements, effectively decomposing visual information into logically ordered shot segments. This process not only enhances the visualization of storytelling but also establishes a coherent link between visual presentation and narrative flow.

However, conventional video editing tasks primarily rely on visual features for shot parsing and feature extraction. While this method effectively captures visual transitions within shots, it often fails to represent higher-level narrative logic and emotional context. For instance, the same shot may convey entirely different emotions and meanings depending on its narrative backdrop—an aspect that purely visual processing struggles to interpret. Moreover, the limitations of visual-only representations lead to a lack of flexibility and interpretability during video editing and re-composition.

To address these challenges, this paper introduces a universal shot representation method based on LLMs, termed L-Storyboard. Inspired by traditional storyboarding techniques, L-Storyboard leverages language descriptions to structurally express discrete shots, explicitly presenting complex shot language and narrative intentions. This language-based representation not only enhances the capture of high-level shot semantics but also aligns with the reasoning and generative capabilities of LLMs, enabling more flexible and efficient video editing tasks. Furthermore, the language modality, as a high-information-density representation, naturally encapsulates the semantic characteristics and editorial intentions of shots, thereby improving the interpretability and extensibility of video editing processes.

\subsection{Generation Process}

The generation process of L-Storyboard consists of four main stages: shot detection, shot language extraction, audiovisual information extraction, and unified information representation.

In the shot detection stage, the system analyzes both user-recorded footage and externally sourced videos. For online videos, boundary detection algorithms \cite{cite35} are applied to segment the continuous video stream into discrete shot fragments. Each fragment is annotated with a timestamp, serving as the basic processing unit for subsequent stages.
In the shot detection stage, the system analyzes both user-recorded footage and externally sourced videos. For online videos, boundary detection algorithms \cite{cite35} are applied to segment the continuous video stream into discrete shot fragments. Each fragment is annotated with a timestamp, serving as the basic processing unit for subsequent stages.

Following shot detection, the system transitions to the shot language analysis stage. This phase focuses on identifying the visual expressions and narrative functions of each shot, guided by cinematographic definitions of shot language. This analysis serves as the foundational representation for L-Storyboard. We employ the ShotTransformer \cite{cite34} as the primary feature extraction model. This model utilizes a dual-stream visual Transformer architecture, which integrates dynamic and static backbones to simultaneously capture motion dynamics and compositional structures within shots. This unified design enables precise parsing of shot language. Notably, in Section \ref{sec:sac}, we further demonstrate shot language analysis methods based on VLMs to compare the effectiveness of various technical approaches.

For the multimodal feature extraction of audiovisual information, we integrate the Minicpm-o \cite{cite39} model, which is deployed locally to parse the multimodal content of videos (Figure \ref{fig:2} illustrates the prompt design used). Minicpm-o is a lightweight, multimodal understanding model capable of efficiently parsing visual and linguistic features. In its quantized version, it requires only 5.7 GB of VRAM for deployment, making it compatible with most consumer-grade GPUs. Compared to solutions reliant on high-performance hardware or cloud-based services, local deployment of Minicpm-o not only reduces hardware costs but also enhances data security. For audio processing, we employ Whisper \cite{cite36}, a high-performance Automatic Speech Recognition (ASR) tool that accurately transcribes spoken content from videos into text, while also attaching precise timestamp information.

\begin{figure}[ht]
 \centering 

\begin{exampleprompt}
\texttt{\textbf{SYSTEM:} You are a model assistant specializing in video shot analysis. Based on the following consecutive frame images, please describe the events or changes happening in the scene. Identify the setting, characters, actions, and movement of objects, and indicate if there is a noticeable time progression or plot development.The description should be coherent, specific, and strive to capture the dynamic information present in the frames.}

\texttt{… (The complete prompt can be found in the Supplementary Material.)}

\end{exampleprompt}
 \caption{The prompt design adopted for describing video content.}
 \label{fig:2}
\end{figure}

\begin{figure*}[ht]
 \centering 
 \includegraphics[width=1.7\columnwidth]{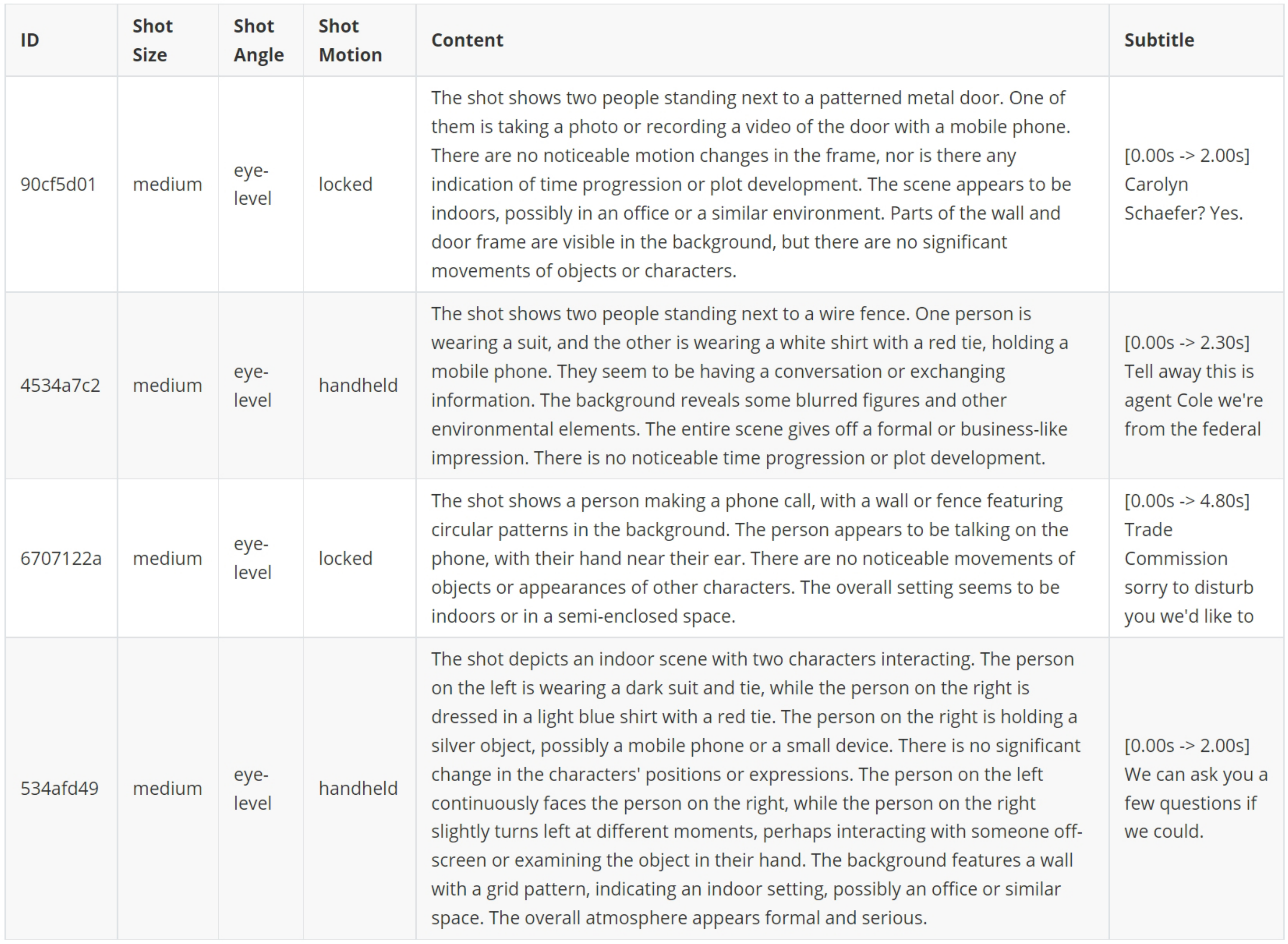}
 \caption{A complete L-Storyboard example, derived from Figure \ref{fig:1} (c).}
 \label{fig:3}
\end{figure*}

Finally, all extracted information is integrated into a unified format and represented in Markdown. As one of the standard training formats for Large Language Models, Markdown offers advantages of being lightweight, structured, and highly compatible. In practical implementation, each shot's visual description, shot language, audio subtitle information, and timestamp are organized into a Markdown table format. As illustrated in Figure \ref{fig:3}, when multiple shot segments are present, the system automatically generates multiple rows within the Markdown table, streamlining the representation process.

\section{LLM-Assisted Video Editing}
\subsection{Convergent Tasks}
In AI-assisted video editing, convergent tasks are primarily concerned with "converging" towards an optimal solution from multiple possibilities. These tasks are typically well-defined with a constrained solution space, making them highly amenable to optimization through efficient reasoning and deep semantic understanding. This study focuses on two representative convergent tasks: Shot Attributes Classification and Next Shot Selection, evaluating their performance and effectiveness under the L-Storyboard-based intermediate representation.

\subsubsection{Shot Attributes Classification} \label{sec:sac}
Shot Attributes Classification (SAC) involves extracting various cinematographically defined features from individual video shots, such as shot scale, shot motion, and shot angle. Cinemetrics \cite{cite37} is a methodological approach in film studies that analyzes shot styles through statistical examination of large datasets of film shots. The Cinemetrics website \cite{cite38} supports the statistical analysis of shot language across numerous films, providing valuable data for computational film studies. Early statistical methods largely relied on manual operations, but the advent of deep learning has significantly enhanced the scalability and automation of shot language analysis.

In deep learning approaches, each shot type is typically treated as an independent task, with classic video analysis neural networks—such as SlowFast \cite{cite40}, R3D \cite{cite41}, MViT \cite{cite42}, and VideoSwin \cite{cite43}—serving as the primary feature extraction models. Research \cite{cite2} has shown that potential shared features exist among different shot types, which can be further leveraged through multi-task learning to improve classification performance.

In our study, we employed various open-source VLMs, including LLAVA \cite{cite44}, MiniCPM-v \cite{cite45}, MiniCPM-o \cite{cite39}, and LLaMA-3.2V \cite{cite46}, to perform SAC across multiple shot types under single-turn dialogue conditions. We also conducted a systematic comparative analysis with traditional deep learning methods. Moreover, we fine-tuned MiniCPM-v using LoRA (Low-Rank Adaptation of Large Language Models) \cite{cite47} with parameters set to rank=32 and alpha=8, effectively enhancing the model's accuracy in shot attributes classification. Figure \ref{fig:4} illustrates the prompt design used for SAC. In addition to task prompts, detailed definitions and candidate options for each shot type are provided as input. \textit{Due to the extended length of the complete prompt, specific details are included in the Supplementary Material, which also applies to subsequent sections.
}
\begin{figure}[ht]
 \centering 

 \begin{exampleprompt}
\texttt{\textbf{SYSTEM:} You are a model assistant focused on video shot analysis, specializing in predicting {shot} based on shot content. Please carefully observe shot details and make predictions according to the following definitions:}

\texttt{… (The complete prompt can be found in the Supplementary Material.) \textbf{/think}}

\end{exampleprompt}
 \caption{The prompt design used in Shot Attributes Classification.}
 \label{fig:4}
\end{figure}

\subsubsection{Next Shot Selection}
Next Shot Selection (NSS) aims to infer the most appropriate subsequent shot based on the semantic content and context of the current shot. This task is fundamental to video editing, yet its implementation via deep learning methods remains challenging. Typically, NSS is modeled through contrastive learning, represented as follows: 

Let the existing shot sequence be $S=\{s_1,s_2,…,s_n\}$, and the candidate shot sequence be $C=\{c_1,c_2,…,c_m\}$. The correct subsequent shot sequence is denoted as the positive sample $S^+=\{s_1,s_2,…,s_n\}$, if and only if $c_i=s_(n+1)$; otherwise, it is considered a negative sample $S^-=\{s_1,s_2,…,c_i \}$, where $c_i \ne s_(n+1)$. The objective is to maximize the similarity of the positive samples and minimize the similarity of the negative samples, thereby optimizing the model's ability to predict reasonable shot sequences.

For the negative sample sampling strategy in contrastive learning, \cite{cite1} proposes two strategies: 1. In-sequence Sampling: For each input sequence, $k-n-1$ incorrect candidate shots are selected from the candidate list as negative samples. 2. Within a batch, all other shot sequences (excluding the correct ones) are considered additional negative samples.

We propose an L-Storyboard-based approach that transforms NSS into a more natural and intuitive process. L-Storyboard relies entirely on textual information as input, effectively mitigating privacy risks associated with personal image exposure, thereby ensuring data security. Additionally, we introduce the CoT (Chain-of-Thought) \cite{cite12} reasoning strategy to better simulate the editorial decision-making process during shot transitions. Figure \ref{fig:5} illustrates the prompt design used for NSS.
\begin{figure}[ht]
 \centering 

  \begin{exampleprompt}
\texttt{\textbf{SYSTEM:} You are an experienced film editing analyst. You will read a storyboard table containing the following information for each shot: ID, shot size, shot angle, shot movement, shot content, and subtitles. Your tasks are:
}

\texttt{… (The complete prompt can be found in the Supplementary Material.) \textbf{/think}}

\end{exampleprompt}
 \caption{The prompt design used in Next Shot Selection.}
 \label{fig:5}
\end{figure}

\subsection{Divergent Tasks Optimization}
Divergent tasks are characterized by a vast solution space with multiple plausible outcomes. These tasks rely heavily on high-level narrative logic and emotional rhythm control, aspects that traditional deep learning methods, which primarily depend on visual features, struggle to handle in multi-path reasoning scenarios. This study focuses on Shot Sequence Ordering as a representative divergent task and proposes an innovative optimization strategy called StoryFlow. By employing multi-temperature generation and convergent selection mechanisms, this approach transforms the divergent nature of the task into a more convergent process, thereby improving accuracy and coherence.

\subsubsection{Shot Sequence Ordering}
Shot Sequence Ordering (SSO) involves reorganizing a collection of unordered discrete shots into coherent narrative segments based on story logic. From an editorial perspective, variations in montage techniques and individual editing styles allow for multiple valid shot sequences. Thus, SSO inherently lacks a single, definitive answer, making it a prototypical divergent task.

Traditional deep learning algorithms often frame shot sequencing as a classification problem, where the number of categories corresponds to all possible permutations \cite{cite1}. However, as the number of shots increases, the permutations grow exponentially, complicating training and interpretability. The use of L-Storyboard partially alleviates these issues by representing outputs as text information rather than high-dimensional vectors. However, our experiments revealed that LLMs exhibit inherent randomness, resulting in uncertainty in output sequences.
To address this, we reference \cite{cite7} and introduce Kendall's Tau Distance as a relative metric—as opposed to the absolute measure of Top-1 Accuracy—to assess the similarity between the predicted sequence and the ground truth. A value of 0 indicates perfect alignment.

Building on this, we further explored Few-shot, Zero-shot, and LoRA \cite{cite47} approaches across various LLMs, including Qwen3 \cite{cite50} and DeepSeek \cite{cite51}. Figure \ref{fig:6} illustrates the prompt design for SSO. However, initial experimental results fell short of expectations, motivating the development of the StoryFlow strategy.

\begin{figure}[ht]
 \centering 
  \begin{exampleprompt}
\texttt{\textbf{SYSTEM:} You are an experienced film editing analyst. You will read a storyboard information table that lists the ID, shot size, shot angle, shot movement, shot content, and subtitles for each shot. Based on this information, infer the actual order in which these shots appear in the story.
}

\texttt{… (The complete prompt can be found in the Supplementary Material.) \textbf{/think}}

\end{exampleprompt}
 \caption{The prompt design used in Shot Sequence Order.}
 \label{fig:6}
\end{figure}

\subsubsection{StoryFlow: Multi-Temperature Divergence and Convergent Narrative Selection}

The output of Shot Sequence Ordering is often unstable; unless both the seed and temperature parameters are fixed, inference results can yield different sequences under identical conditions. However, extensive experimentation revealed that while some randomness exists, there is also an underlying logic to the outputs. Results are confined within a relatively stable solution space, with experimental analysis indicating an average of 2.34 reasonable orderings for three-shot sequencing tasks \cite{cite1}.

\begin{figure*}[ht]
 \centering 
 \includegraphics[width=2\columnwidth]{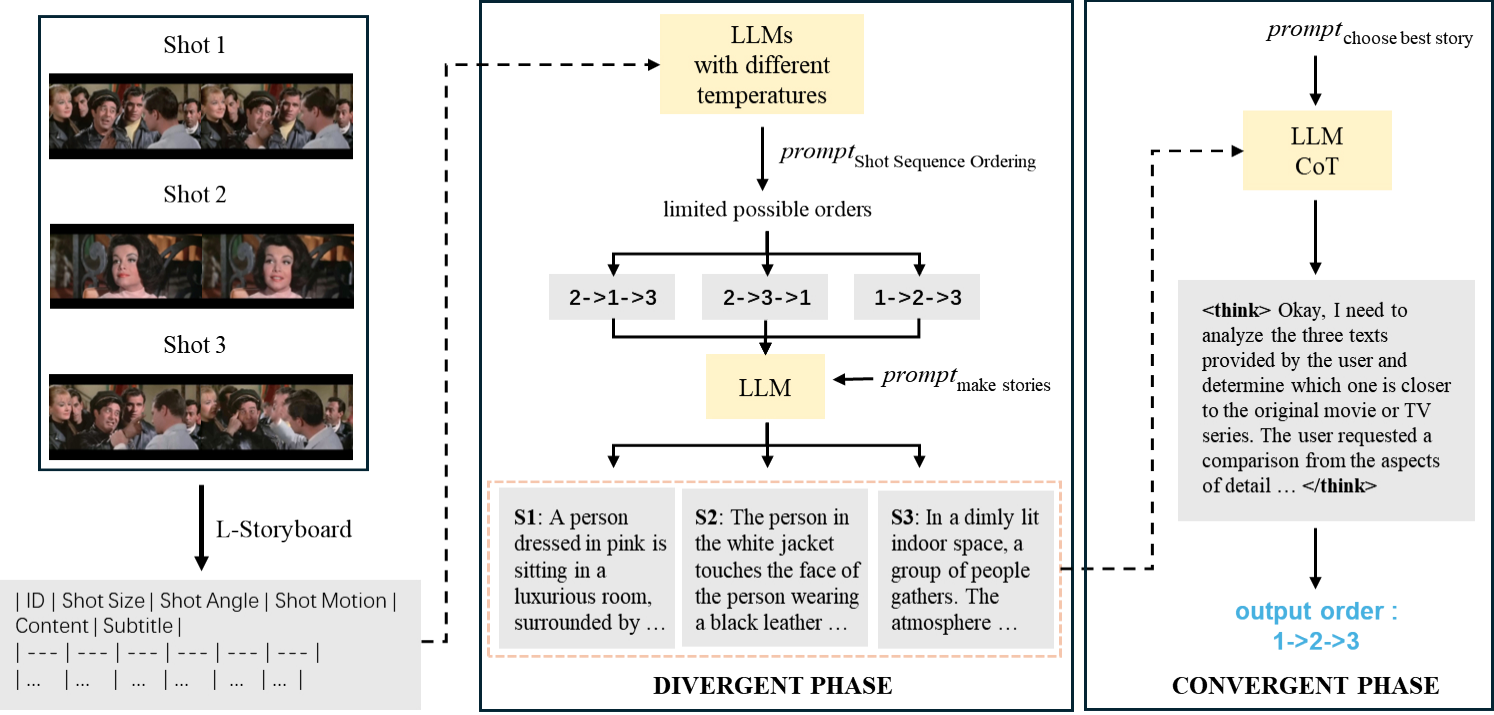}
 \caption{Our proposed StoryFlow strategy consists of two phases: Divergent Phase and Convergent Phase. In the Divergent Phase, different temperature settings are used to generate multiple possible shot sequences, which are then structured into complete story segments based on L-Storyboard. In the Convergent Phase, the model selects the version that best aligns with narrative logic, producing a more coherent and logically sound shot sequence.}
 \label{fig:7}
\end{figure*}

Building on this characteristic, we propose the StoryFlow optimization strategy. As illustrated in Figure \ref{fig:7}, StoryFlow is structured into two distinct phases: the Divergent Phase and the Convergent Phase.
In the Divergent Phase, the LLM is configured with various temperature settings to represent different levels of creativity, ranging from 0 to 2 with a step size of 0.2. Using the prompt design depicted in Figure \ref{fig:6}, the model generates a series of plausible shot sequence candidates. These sequences are assumed to be valid solutions and, guided by the L-Storyboard descriptions, are organized into independent and complete story segments.

In the Convergent Phase, the diverse story segments produced in the Divergent Phase serve as inputs, effectively transforming the task into a convergent problem. The LLM is then directed to select the version that best aligns with cinematic logic or narrative rhythm, aiming to emulate the storytelling style of professional film and television editing. Figure \ref{fig:8} illustrates the prompt designs employed in both the Divergent and Convergent Phases, respectively.

\begin{figure}[ht]
 \centering 

   \begin{exampleprompt}
\texttt{\textbf{DIVERGENT PHASE: } You are a film narrative script expert, skilled at writing coherent stories based on shot storyboards and shot sequences. The user will provide the storyboard and the shot sequence, and you need to strictly follow the order, integrating each shot's content and subtitles into the story, creating a complete yet concise narrative.The output should only contain the story itself, with no explanations, annotations, analysis, or formatting language.
}

\end{exampleprompt}
  \begin{exampleprompt}
\texttt{\textbf{CONVERGENT PHASE: } You are an expert in film and television plot analysis, capable of assessing how closely a provided text aligns with the actual storyline of a movie or TV series. Your task is to compare the given text with the original work based on details, language style, and plot development, and determine which text is more faithful to the source material. Provide a clear and concise conclusion with a reasoned explanation.
}

\end{exampleprompt}
 \caption{The $prompt_{make \ stories}$ and $prompt_{choose \ best \ story}$ used in the StoryFlow strategy.}
 \label{fig:8}
\end{figure}

The core concept of StoryFlow is to transform the inherently divergent nature of shot sequencing into a convergent process. Its multi-stage processing mechanism allows for enhanced flexibility and adaptability, such as adjusting the Convergent Phase prompt to modify output styles. Notably, StoryFlow achieves this without requiring additional model training, making it easily integrable with existing LLMs. Experimental results in Section \ref{sec:result_sso} further validate that this method significantly improves the performance of shot sequencing tasks, as measured by relative metrics.

\section{ Experimental Results}

\subsection{Implement Details}
\paragraph{Dataset Usage} To evaluate the effectiveness of the L-Storyboard method, we primarily selected the AVE dataset \cite{cite1} as the experimental foundation. This dataset includes comprehensive annotations, authoritative baselines, and human benchmarks for certain tasks, making it highly suitable as a benchmark for this study. Additionally, for the Next Shot Selection task, we introduced the ActivityNet100 \cite{cite52} dataset, which more accurately reflects the characteristics of web-based video distribution. This allowed us to evaluate the model's performance in larger-scale and more diverse video scenarios.

For dataset partitioning, we strictly adhered to the standards outlined in the original papers \cite{cite1}\cite{cite52}. During the LoRA fine-tuning phase for Shot Attributes Classification, the training and testing sets were entirely distinct. For the Next Shot Selection task, following \cite{cite1}, we sequentially sampled 9 shots from each video segment, designating the first four as the known sequence and the remaining five as candidate shots. For Shot Sequence Ordering, three discrete shots were selected from each segment for reordering.

\paragraph{Evaluation Metrics} In the Shot Attributes Classification task, we employed Top-1 Accuracy and Macro-F1 as evaluation metrics, while also tracking each model's parameter size to facilitate comparisons in performance and computational complexity. For Next Shot Selection, accuracy served as the primary evaluation metric. In the Shot Sequence Ordering task, we adopted both Top-1 Accuracy and Mean Kendall's Tau Distance (Mean KTD) as standards for assessment.

\paragraph{Parameter Settings} In all experiments outside of StoryFlow, the temperature parameter for all models was set to 0 to ensure stable and consistent inference results. Additionally, in Next Shot Selection, the Video representation in L-Storyboard excluded subtitle information. For models with parameter sizes exceeding 100B, online services were utilized. All other models were deployed using the q4\_k\_m quantized version from Ollama \cite{cite53}. 

\subsection{Results}

\subsubsection{Shot Attributes Classification}

Table \ref{tab:1} presents the overall performance of various methods in the Shot Attributes Classification task. Experimental results indicate that, compared to traditional deep-learning-based video processing methods, most VLMs achieve strong classification performance without fine-tuning. Notably, MiniCPM-v \cite{cite45}, after LoRA fine-tuning, improved Macro-F1 scores for Shot Size and Shot Angle by over 10\% compared to the previously leading model, ShotTransformer \cite{cite34}. This enhancement partially addresses the long-tail distribution issues in traditional deep learning methods that often result in lower Macro-F1 scores.
However, due to VLMs' relatively limited capacity for temporal information processing \cite{cite59}, their performance in Shot Motion, which relies on temporal reasoning, is considerably weaker compared to Video Transformer-based methods.

\begin{table*}[ht]
\centering
\caption{Overall Performance of Shot Attributes Classification}
\label{tab:1}
\resizebox{0.9\textwidth}{!}{
\begin{tabular}{lccccccc}
\toprule
                           & \multicolumn{2}{c}{Shot Size} & \multicolumn{2}{c}{Shot Angle} & \multicolumn{2}{c}{Shot Motion} & \multicolumn{1}{l}{} \\ \midrule
                           
Method                     & Accuracy      & Macro-F1      & Accuracy       & Macro-F1      & Accuracy       & Macro-F1       & Params               \\
R3D \cite{cite41}                   & 67.45         & 24.97         & 85.37          & 19.05         & 70.18          & 29.02          & 33.18M               \\
SlowFast \cite{cite40}              & 66.10         & 20.38         & 81.13          & 19.72         & 68.17          & 29.40          & 33.57M               \\
MViT \cite{cite42}                  & 73.54         & 41.27         & 87.19          & 33.04         & 71.63          & 36.62          & 34.11M               \\
ViViT \cite{cite54}                 & 69.30         & 28.95         & 85.44          & 18.43         & 70.11          & 28.91          & 114.59M              \\
CLIP \cite{cite55}                  & 51.30         & ——            & 54.90          & ——            & 37.60          & ——             & 120.54M              \\
Long-range multimodal \cite{cite15} & 67.40         & ——            & 57.70          & ——            & 46.10          & ——             & ——                   \\
ShotTransformer \cite{cite34}       & 73.39         & 44.61         & 86.91          & 34.72         & 70.85          & 38.11          & 460.69M              \\
LLAVA \cite{cite44}                 & 68.76         & 21.73         & 74.97          & 24.32         & 47.39          & 12.99          & 11B                  \\
MiniCPM-v \cite{cite45}             & 55.27         & 26.64         & 83.22          & 25.20         & 46.87          & 13.80          & 8B                   \\
MiniCPM-o \cite{cite39}             & 75.41         & 23.23         & 84.86          & 20.07         & 47.48          & 12.56          & 8B                   \\
LLama-3.2 vision \cite{cite46}      & 52.68         & 24.47         & 72.73          & 22.72         & 47.28          & 12.56          & 8B                   \\
Minicpm-v LoRA \cite{cite47}        & 80.95         & 49.23         & 87.90          & 45.16         & 61.59          & 26.38          & 8B                  \\ \bottomrule
\end{tabular}}
\end{table*}

From a parameter perspective, the two best-performing models, ShotTransformer and MiniCPM-v LoRA \cite{cite47}, exhibit substantial differences in computational cost. The parameter size of MiniCPM-v LoRA is 17 times larger than that of ShotTransformer, and even after q4\_k\_m quantization, its model size remains 5 times that of ShotTransformer. This suggests that while significant performance gains are achieved, they come with considerable computational overhead. Overall, Video Transformer-based methods demonstrate superior capabilities in modeling temporal information, whereas VLMs excel in the classification of static shots. Thus, in practical applications, model selection should be guided by the task's dependency on temporal reasoning to strike a balance between performance and resource consumption.

\subsubsection{Next Shot Selection}

Table \ref{tab:2} summarizes the performance comparison of various methods in the Next Shot Selection task, evaluated on both the AVE and ActivityNet100 datasets. Our experiments compared the pure visual (Video) and audiovisual (Audio + Video) modes in terms of accuracy. Under traditional Contrastive Learning methods, the AVE dataset achieved 37.5\% / 38.7\% accuracy for the visual mode and 41.0\% / 41.4\% for the audiovisual mode.

\begin{table*}[ht]
\centering
\caption{Overall Performance of Next Shot Selection}
\label{tab:2}
\resizebox{0.7\textwidth}{!}{
\begin{tabular}{lcccc}
\toprule
                                        & \multicolumn{2}{c}{AVE} & \multicolumn{2}{c}{ActivityNet 100} \\ \midrule
                                        & Video  & Audio + Video  & Video        & Audio + Video        \\
Random                                  & 20.0   & 20.0           & 20.0         & 20.0                 \\
Contrastive learning (in-sequence) \cite{cite56} & 37.5   & 38.7           & ——           & ——                   \\
Contrastive learning (in-batch) \cite{cite56}    & 41.0   & 41.4           & ——           & ——                   \\
Qwen3:1.7B \cite{cite50}                         & 28.23  & 31.25          & 27.42        & 30.97                \\
Qwen3:4B \cite{cite50}                           & 24.92  & 35.05          & 32.14        & 29.56                \\
Qwen3:8B \cite{cite50}                           & 41.78  & 47.51          & 40.75        & 47.03                \\
Qwen3:14B \cite{cite50}                          & 42.48  & 46.19          & 44.00        & 41.99                \\
Qwen3:32B \cite{cite50}                          & 48.56  & 51.63          & 44.04        & 47.90                \\
DeepSeek-R1:671B \cite{cite51}                   & 43.60  & 45.98          & ——           & ——                   \\
Qwen3-235B-a22b \cite{cite50}                    & 42.74  & 47.78          & 43.29        & 45.67  \\ \bottomrule             
\end{tabular}
}
\end{table*}

After incorporating L-Storyboard, the performance of different versions of Qwen3 \cite{cite50} improved progressively with increased parameter scales. On the AVE dataset, Qwen3:1.7B achieved 28.23\% accuracy in the Video mode and 31.25\% in the Audio + Video mode. Meanwhile, Qwen3:32B saw its accuracy rise to 48.56\% and 51.63\%, respectively, significantly surpassing traditional contrastive learning methods. This suggests that larger parameter scales enhance the model's understanding of logical relationships and contextual coherence between shots.

A similar trend was observed on the ActivityNet100 dataset, where Qwen3:32B reached 47.90\% in the Audio + Video mode. However, when models were further scaled up to DeepSeek-R1:671B and Qwen3-235B-a22b, performance unexpectedly declined. This suggests that increasing parameter scales does not always yield linear performance improvements and may introduce issues like overfitting or diminished temporal reasoning efficiency.

\subsubsection{Shot Sequence Ordering}\label{sec:result_sso}

Table \ref{tab:3} displays the overall performance of different methods in the Shot Sequence Ordering task, including traditional deep learning models, various configurations of LLMs (Zero-Shot, Few-Shot), and our proposed StoryFlow strategy.

\begin{table*}[ht]
\centering
\caption{Overall Performance of Shot Sequence Ordering}
\label{tab:3}
\resizebox{1\textwidth}{!}{
\begin{tabular}{lcclclccc}
\toprule
Method                & \multicolumn{1}{l}{} & \multicolumn{1}{l}{}       & \multicolumn{1}{c}{Method} & \multicolumn{3}{c}{Zero-Shot}                      & \multicolumn{2}{c}{Few-Shot} \\
                      & Top-1 Accuracy       & Mean KTD                   &                            & \multicolumn{2}{c}{Top-1 Accuracy} & Mean KTD      & Top-1 Accuracy↑  & Mean KTD  \\ \midrule
Random                & 16.60                & \multicolumn{1}{c|}{1.500} & DeepSeek R1:7B \cite{cite51}        & \multicolumn{2}{c}{18.32}          & 1.427         & 17.45            & 1.509     \\
Human                 & 39.90                & \multicolumn{1}{c|}{——}    & Qwen3:1.7B \cite{cite50}            & \multicolumn{2}{c}{16.49}          & 1.455         & 18.52            & 1.458     \\
R3D early fusion \cite{cite41} & 25.70                & \multicolumn{1}{c|}{——}    & Qwen3:4B \cite{cite50}              & \multicolumn{2}{c}{19.30}          & 1.407         & 18.87            & 1.493     \\
R3D late fusion \cite{cite41}  & 21.50                & \multicolumn{1}{c|}{——}    & Qwen3:8B \cite{cite50}              & \multicolumn{2}{c}{18.92}          & 1.456         & 16.90            & 1.485     \\
SlowFast \cite{cite40}         & 25.29                & \multicolumn{1}{c|}{1.474} & Qwen3:14B \cite{cite50}             & \multicolumn{2}{c}{18.09}          & 1.456         & 17.60            & 1.493     \\
R3D \cite{cite41}              & 25.44                & \multicolumn{1}{c|}{1.469} & Qwen3:32B \cite{cite50}             & \multicolumn{2}{c}{20.98}          & 1.409         & 20.78            & 1.405     \\
MViT \cite{cite42}             & 32.04                & \multicolumn{1}{c|}{1.345} & Qwen2.5:7B \cite{cite50}            & \multicolumn{2}{c}{16.88}          & 1.471         & 18.55            & 1.444     \\
VideoSwin \cite{cite57}        & 27.21                & \multicolumn{1}{c|}{1.474} & Qwen3:4B LoRA \cite{cite47}         & \multicolumn{2}{c}{17.07}          & 1.478         & ——               & ——        \\
VideoMAE \cite{cite58}         & 25.29                & \multicolumn{1}{c|}{1.474} & Qwen3:8B LoRA \cite{cite47}         & \multicolumn{2}{c}{18.13}          & 1.475         & ——               & ——        \\
\textbf{StoryFlow}             & 29.82                & \multicolumn{1}{c|}{\textbf{1.205}} & Qwen3-235b-a22b \cite{cite50}       & \multicolumn{2}{c}{20.09}          & 1.463         & ——               & ——     \\  \bottomrule
\end{tabular}
}
\end{table*}
Among traditional methods, MViT demonstrated the strongest ordering capability, achieving a Top-1 Accuracy of 32.04\% and an average Kendall's Tau Distance (Mean KTD) of 1.345, significantly outperforming other baseline models.

For LLM-based methods, we evaluated both Zero-Shot and Few-Shot scenarios. In the Zero-Shot setting, Qwen3:32B exhibited the best performance, with a Top-1 Accuracy of 20.98\% and a Mean KTD of 1.409, indicating that large language models maintain robust reasoning abilities even without specific task fine-tuning. In the Few-Shot setting, Qwen3:32B continued to lead, with its Top-1 Accuracy improving to 20.78\% and its Mean KTD reducing to 1.405. However, performance declined after LoRA fine-tuning, suggesting that fine-tuning strategies may not significantly benefit shot ordering tasks and might even introduce inference biases.

In contrast, our proposed StoryFlow strategy achieved a Top-1 Accuracy of 29.82\%. Although slightly lower than MViT, it outperformed all other methods in Mean KTD, with a score of 1.205, indicating superior coherence and logical consistency in shot ordering. This improvement is primarily attributed to the Divergent Phase and Convergent Phase design of StoryFlow, which effectively reduces the uncertainty inherent in divergent tasks through multi-temperature exploration and convergent selection.

A further comparison with the Few-Shot results of Qwen3:32B reveals that while their Top-1 Accuracy values are comparable, Qwen3:32B demonstrates a significantly higher Mean KTD, indicating deficiencies in fine-grained ordering logic. Conversely, StoryFlow, with its multi-path reasoning and aggregation strategy, effectively reduces sequencing errors, demonstrating enhanced reliability and logical alignment in complex video narratives.

\subsection{Discussion}
This study introduces LLM-Assisted Video Editing, systematically integrating LLMs into the domain of video editing for the first time and presenting a novel framework. At the core of this framework is L-Storyboard, a method that transforms traditional visual representations into language-based descriptions, mapping the visual information of discrete shots into structured natural language expressions. This transformation not only enhances the interpretability of shot content but also addresses long-standing data privacy concerns in video editing. Unlike traditional visual feature extraction and cloud-based processing, L-Storyboard's language-based representation is processed entirely locally, eliminating the need to upload original videos or images and significantly reducing the risk of data leakage. Moreover, most models employed in our experiments support local deployment and do not require high-performance cloud computing environments. This design enables efficient inference and editing on personal devices, greatly lowering the barrier to entry for advanced video editing technologies.

Unlike conventional deep learning-based approaches, LLMs facilitate high-level reasoning and semantic inference through language modalities once the connection is established. This capability substantially improves both the understanding and execution efficiency of complex video editing tasks. Furthermore, we introduced a systematic categorization of video editing tasks into Convergent Tasks and Divergent Tasks, enabling a more structured and optimized approach for different editing requirements. Convergent tasks—such as Shot Attributes Classification and Next Shot Selection—demonstrated superior performance compared to traditional deep learning methods, leveraging LLMs' language reasoning abilities without requiring fine-tuning.

For divergent tasks, which are traditionally challenging for conventional methods due to their multiple plausible solutions, our proposed StoryFlow strategy showed significant advantages. Divergent tasks inherently involve multi-path reasoning, which standard deep learning approaches struggle to handle effectively. StoryFlow's dual-phase design successfully decomposes this complexity into two phases: Divergent Exploration and Convergent Selection. In the Divergent Phase, LLM-based reasoning is performed at varying temperature settings to generate multiple potential shot sequences. In the Convergent Phase, logical reasoning is applied to select the sequence that best aligns with narrative logic. This two-stage design not only mitigates the high uncertainty associated with multi-path reasoning but also significantly improves the coherence and logical consistency of the final sequencing. Experimental results demonstrate that, compared to traditional Video Transformers and Few-Shot LLM methods, StoryFlow achieves superior Mean Kendall's Tau Distance (Mean KTD) and Top-1 Accuracy in shot ordering tasks.

Moreover, because the reasoning mechanisms of L-Storyboard and StoryFlow are entirely Prompt-driven, the architecture is highly flexible and scalable. Adapting to new video editing tasks only requires modifications to the Prompt design, without changes to the underlying architecture. This adaptability opens avenues for future applications, including multi-shot interactions, complex plot structuring, and even cross-temporal video narrative reconstruction.

\paragraph{Limitations}
Although this study demonstrates the significant potential of LLMs in video editing tasks, several limitations remain that warrant attention.

First, during the transformation of visual information into language descriptions, some degree of information loss is inevitable. This limitation is particularly evident in scenarios involving complex visual dynamics, subtle emotional expressions, and multi-agent interactions, where purely text-based descriptions may fail to capture intricate details, leading to inference biases. For example, in sequences with fast-moving shots or complex backgrounds, the L-Storyboard representation may struggle to accurately convey fine-grained visual details, potentially affecting the quality of shot sequencing.

Second, the current StoryFlow strategy has been primarily validated on Shot Sequence Ordering, a specific divergent task. Its effectiveness in addressing more complex divergent editing tasks—such as long-sequence plot restructuring and non-linear story construction—remains untested. While the dual-phase design of StoryFlow enhances coherence and logical consistency in shot ordering, its applicability to long-span shot reasoning has yet to be empirically verified.

Finally, the diversity of video editing tasks suggests that a single reasoning approach may not be sufficient to address all requirements. Tasks like narrative editing, trailer generation, and dynamic shot switching demand more granular reasoning and a deeper understanding of contextual relationships. Although L-Storyboard and StoryFlow exhibit promising scalability, future research should explore deeper integration with LLMs for a broader range of tasks, particularly focusing on capturing multi-agent interactions and managing complex temporal structures. Addressing these challenges will be a critical direction for future exploration.

\section{Conclusion}
This study presents an LLM-Assisted Video Editing framework, introducing the reasoning and generative capabilities of Large Language Models into the domain of video editing for the first time. Through the innovative design of L-Storyboard, we transform discrete video shots into structured language descriptions, significantly enhancing both the interpretability and privacy protection of video editing tasks. The local deployment of L-Storyboard reduces the complexity of visual information processing and mitigates the risks of data privacy leaks, while simultaneously demonstrating strong scalability and adaptability.

From a task categorization perspective, we systematically classified video editing into Convergent Tasks and Divergent Tasks. For Convergent Tasks (e.g., Shot Attributes Classification and Next Shot Selection), LLM's language reasoning capabilities delivered superior accuracy and interpretability compared to traditional deep learning methods. For more complex Divergent Tasks (e.g., Shot Sequence Ordering), the StoryFlow strategy effectively improved logical coherence and ordering accuracy through its dual-phase reasoning mechanism. Experimental results confirm that the proposed methods consistently outperform traditional approaches across various video editing tasks. L-Storyboard, mapped with LLMs, outperformed conventional deep learning models in Shot Classification and Next Shot Selection, while StoryFlow demonstrated greater logical consistency and accuracy in Shot Sequence Ordering.

Overall, the LLM-Assisted Video Editing framework introduced in this study, along with the innovative designs of L-Storyboard and StoryFlow, provides a novel solution for video editing tasks, highlighting the vast potential of Large Language Models in complex video understanding and reasoning. Looking forward, we anticipate exploring more intelligent and scalable solutions for advanced video editing tasks based on this framework, further pushing the boundaries of AI-driven video production and post-production capabilities.

\cleardoublepage
\twocolumn[
    \begin{center}
        {\LARGE Supplementary Material\par}  % 大字体、加粗
        \vspace{0.5em} % 间距
       
    \end{center}
]

\section*{Complete Prompt of Shot Attributes Classification}

\texttt{You are a model assistant focused on video shot analysis, specializing in predicting {shot} based on shot content. Please carefully observe shot details and make predictions according to the following definitions:}

\texttt{\#\# shot-size}

\texttt{Determine the shot size based on the amount of subject and surrounding environment shown. Choose one: 'medium', 'wide', 'close-up', 'extreme-wide', 'extreme-close-up'.}

\texttt{Definition: }

\texttt{shot-size is defined as how much of the setting or subject is displayed within a given shot. shot-size has five categories: 
}

\texttt{1) `extreme-wide` shots barely show the subject and the shot's main focus is the subject's surrounding; 
}

\texttt{2) `wide` shots, also known as long shot, show the entire subject and their relation to the surrounding environment; 
}

\texttt{3) `medium` shots depict the subject approximately from the waist up emphasizing both the subject and their surrounding; 
}

\texttt{4) `close-up` shots are taken at a close range intended to show greater detail to the viewer; 
}

\texttt{5) `extreme-close-up` shots frame a subject very closely where the outer portions of the subject are often cut off by the frame's edges.
}

\texttt{\#\# shot-angle
}

\texttt{Determine the shooting angle, which is the camera position relative to the subject. Choose one: 'eye-level', 'high-angle', 'low-angle', 'overhead', 'aerial'.
}

\texttt{Definition: 
}
\texttt{shot-angle is the location where the camera is placed to take a shot. shot-angle has five categories: 
}

\texttt{1) `aerial` shot is captured from an elevated vantage point; 
}

\texttt{2) `overhead` shot is when the camera is placed directly above the subject; 
}

\texttt{3) `eye-level` shot is a shot where the camera is positioned directly at the subject's eye level; 
}

\texttt{4) `high-angle` shot is when the camera points down on the subject from above; 
}

\texttt{5) `low-angle` shot is when the camera is positioned below the eye level and looks up at the subject.
}

\texttt{\#\# shot-motion}

\texttt{Analyze the camera movement in the shot. Choose one: 'locked', 'handheld', 'tilt', 'zoom', 'pan'
}

\texttt{Definition: 
}

\texttt{shot-motion is defined as the movement of the camera when taking a shot. shot-motion has five categories: 
}

\texttt{1) `pan` shot is when the camera is moving horizontally while its base remains in a fixed position; 
}

\texttt{2) `tilt` shot is when the camera moves vertically up or down with its base fixated to a certain point; 
}

\texttt{3) `locked` is taken without shifting the position of the camera; 
}

\texttt{4) `zoom` shot is when the camera moves forward and backward adding depth to a scene; 
}

\texttt{5) `handheld` shot is taken with the camera being supported only by the operator's hands and shoulder.
}

\section*{Complete Prompt of Next Shot Selection}
\texttt{You are an experienced film editing analyst. You will read a storyboard table containing the following information for each shot: ID, shot size, camera angle, camera movement, shot content, and subtitles.}

\texttt{Your tasks are:
}

\texttt{1.	Read the "Sequential Shots" information to understand the scene, rhythm, and plot logic;
}

\texttt{2.	Read the "Candidate Shots" information;
}

\texttt{3.	Based on the following criteria, determine which candidate shot is most likely to be the next shot:
}

\texttt{· Spatial continuity: Whether the scene or background is consistent or naturally connected;
}

\texttt{· Continuity of character actions and eye lines;
}

\texttt{· Logical coherence of plot and dialogue;
}

\texttt{· Reasonableness of shot language rhythm;
}

\texttt{· Stylistic consistency (coordination of shot size and movement style);
}

\section*{Complete Prompt of Shot Sequence Order}
\texttt{You are an experienced film editing analyst. You will read a storyboard information table that lists the ID, shot size, camera angle, camera movement, shot content, and subtitles for each shot. Based on this information, infer the actual order in which these shots appear in the story.
}

\texttt{Your task is to comprehensively analyze the visual content, character actions, dialogue sequence, and possible scene clues to determine which shot should appear first, which should follow, and which should come last.
}

\texttt{Please output a string in the format "Shot ID 1->Shot ID 2->Shot ID 3..." (using "->" to connect the shot IDs).
}

\begin{thebibliography}{10}
\providecommand{\url}[1]{#1}
\csname url@samestyle\endcsname
\providecommand{\newblock}{\relax}
\providecommand{\bibinfo}[2]{#2}
\providecommand{\BIBentrySTDinterwordspacing}{\spaceskip=0pt\relax}
\providecommand{\BIBentryALTinterwordstretchfactor}{4}
\providecommand{\BIBentryALTinterwordspacing}{\spaceskip=\fontdimen2\font plus
\BIBentryALTinterwordstretchfactor\fontdimen3\font minus \fontdimen4\font\relax}
\providecommand{\BIBforeignlanguage}[2]{{%
\expandafter\ifx\csname l@#1\endcsname\relax
\typeout{** WARNING: IEEEtran.bst: No hyphenation pattern has been}%
\typeout{** loaded for the language `#1'. Using the pattern for}%
\typeout{** the default language instead.}%
\else
\language=\csname l@#1\endcsname
\fi
#2}}
\providecommand{\BIBdecl}{\relax}
\BIBdecl

\bibitem{cite9}
Y.~Ge, W.~Hua, K.~Mei, j.~ji, J.~Tan, S.~Xu, Z.~Li, and Y.~Zhang, ``Openagi: When llm meets domain experts,'' in \emph{Advances in Neural Information Processing Systems}, A.~Oh, T.~Naumann, A.~Globerson, K.~Saenko, M.~Hardt, and S.~Levine, Eds., vol.~36.\hskip 1em plus 0.5em minus 0.4em\relax Curran Associates, Inc., 2023, pp. 5539--5568.

\bibitem{cite10}
Z.~Shen, ``Llm with tools: A survey,'' \emph{arXiv preprint arXiv:2409.18807}, 2024.

\bibitem{cite32}
J.~Zhang, J.~Huang, S.~Jin, and S.~Lu, ``Vision-language models for vision tasks: A survey,'' \emph{IEEE Transactions on Pattern Analysis and Machine Intelligence}, vol.~46, no.~8, pp. 5625--5644, 2024.

\bibitem{cite33}
K.~Zhou, J.~Yang, C.~C. Loy, and Z.~Liu, ``Learning to prompt for vision-language models,'' \emph{International Journal of Computer Vision}, vol. 130, no.~9, pp. 2337--2348, 2022.

\bibitem{cite4}
Y.~Huang, S.~Lv, K.-K. Tseng, P.-J. Tseng, X.~Xie, and R.~F.-Y.~L. and, ``Recent advances in artificial intelligence for video production system,'' \emph{Enterprise Information Systems}, vol.~17, no.~11, p. 2246188, 2023.

\bibitem{cite5}
A.~Rao, J.~Wang, L.~Xu, X.~Jiang, Q.~Huang, B.~Zhou, and D.~Lin, ``A unified framework for shot type classification based on subject centric lens,'' in \emph{Computer Vision--ECCV 2020: 16th European Conference, Glasgow, UK, August 23--28, 2020, Proceedings, Part XI 16}.\hskip 1em plus 0.5em minus 0.4em\relax Springer, 2020, pp. 17--34.

\bibitem{cite6}
L.~YZBGN and Y.~YZQWZ, ``Representation learning of next shot selection for vlog editing,'' in \emph{CVEU Workshop at ICCV}, vol.~3, 2023.

\bibitem{cite7}
Y.~Li, H.~Xu, and F.~Tian, ``Shot sequence ordering for video editing: Benchmarks, metrics, and cinematology-inspired computing methods,'' \emph{arXiv preprint arXiv:2503.17975}, 2025.

\bibitem{cite8}
R.~R. Selvaraju, M.~Cogswell, A.~Das, R.~Vedantam, D.~Parikh, and D.~Batra, ``Grad-cam: Visual explanations from deep networks via gradient-based localization,'' in \emph{Proceedings of the IEEE international conference on computer vision}, 2017, pp. 618--626.

\bibitem{cite11}
R.~Spottiswoode, \emph{A grammar of the film: An analysis of film technique}.\hskip 1em plus 0.5em minus 0.4em\relax Univ of California Press, 1969.

\bibitem{cite1}
D.~M. Argaw, F.~C. Heilbron, J.-Y. Lee, M.~Woodson, and I.~S. Kweon, ``The anatomy of video editing: A dataset and benchmark suite for ai-assisted video editing,'' in \emph{Computer Vision -- ECCV 2022}, S.~Avidan, G.~Brostow, M.~Ciss{\'e}, G.~M. Farinella, and T.~Hassner, Eds.\hskip 1em plus 0.5em minus 0.4em\relax Cham: Springer Nature Switzerland, 2022, pp. 201--218.

\bibitem{cite2}
Y.~Li, H.~Xu, F.~Cai, and F.~Tian, ``Improving ai-assisted video editing: Optimized footage analysis through multi-task learning,'' \emph{Neurocomputing}, vol. 609, p. 128485, 2024.

\bibitem{cite3}
F.~Lu, Y.~Li, and F.~Tian, ``Enhancing interpretability in film shot analysis through continuous shot integration and saliency maps,'' \emph{Signal, Image and Video Processing}, vol.~19, no.~4, p. 286, 2025.

\bibitem{cite12}
J.~Wei, X.~Wang, D.~Schuurmans, M.~Bosma, b.~ichter, F.~Xia, E.~Chi, Q.~V. Le, and D.~Zhou, ``Chain-of-thought prompting elicits reasoning in large language models,'' in \emph{Advances in Neural Information Processing Systems}, S.~Koyejo, S.~Mohamed, A.~Agarwal, D.~Belgrave, K.~Cho, and A.~Oh, Eds., vol.~35.\hskip 1em plus 0.5em minus 0.4em\relax Curran Associates, Inc., 2022, pp. 24\,824--24\,837.

\bibitem{cite15}
D.~M. Argaw, J.-Y. Lee, M.~Woodson, I.~S. Kweon, and F.~C. Heilbron, ``Long-range multimodal pretraining for movie understanding,'' in \emph{Proceedings of the IEEE/CVF International Conference on Computer Vision}, 2023, pp. 13\,392--13\,403.

\bibitem{cite13}
Q.~Huang, W.~Liu, and D.~Lin, ``Person search in videos with one portrait through visual and temporal links,'' in \emph{Proceedings of the European conference on computer vision (ECCV)}, 2018, pp. 425--441.

\bibitem{cite14}
Y.~Xiong, Q.~Huang, L.~Guo, H.~Zhou, B.~Zhou, and D.~Lin, ``A graph-based framework to bridge movies and synopses,'' in \emph{Proceedings of the IEEE/CVF International Conference on Computer Vision}, 2019, pp. 4592--4601.

\bibitem{cite16}
F.~Lu, Y.~Li, and F.~Tian, ``Exploring challenge and explainable shot type classification using sam-guided approaches,'' \emph{Signal, Image and Video Processing}, vol.~18, no.~3, pp. 2533--2542, 2024.

\bibitem{cite17}
X.~Jiang, L.~Jin, A.~Rao, L.~Xu, and D.~Lin, ``Jointly learning the attributes and composition of shots for boundary detection in videos,'' \emph{IEEE Transactions on Multimedia}, vol.~24, pp. 3049--3059, 2021.

\bibitem{cite18}
E.~Apostolidis, E.~Adamantidou, A.~I. Metsai, V.~Mezaris, and I.~Patras, ``Video summarization using deep neural networks: A survey,'' \emph{Proceedings of the IEEE}, vol. 109, no.~11, pp. 1838--1863, 2021.

\bibitem{cite20}
D.~M. Argaw, M.~Soldan, A.~Pardo, C.~Zhao, F.~C. Heilbron, J.~S. Chung, and B.~Ghanem, ``Towards automated movie trailer generation,'' in \emph{Proceedings of the IEEE/CVF Conference on Computer Vision and Pattern Recognition}, 2024, pp. 7445--7454.

\bibitem{cite21}
M.~Wang, G.-W. Yang, S.-M. Hu, S.-T. Yau, A.~Shamir \emph{et~al.}, ``Write-a-video: computational video montage from themed text.'' \emph{ACM Trans. Graph.}, vol.~38, no.~6, pp. 177--1, 2019.

\bibitem{cite22}
A.~Truong, F.~Berthouzoz, W.~Li, and M.~Agrawala, ``Quickcut: An interactive tool for editing narrated video,'' in \emph{Proceedings of the 29th Annual Symposium on User Interface Software and Technology}, ser. UIST '16.\hskip 1em plus 0.5em minus 0.4em\relax New York, NY, USA: Association for Computing Machinery, 2016, p. 497–507.

\bibitem{cite19}
P.~Papalampidi, F.~Keller, and M.~Lapata, ``Finding the right moment: Human-assisted trailer creation via task composition,'' \emph{IEEE Transactions on Pattern Analysis and Machine Intelligence}, vol.~46, no.~1, pp. 292--304, 2024.

\bibitem{cite23}
A.~Hurst, A.~Lerer, A.~P. Goucher, A.~Perelman, A.~Ramesh, A.~Clark, A.~Ostrow, A.~Welihinda, A.~Hayes, A.~Radford \emph{et~al.}, ``Gpt-4o system card,'' \emph{arXiv preprint arXiv:2410.21276}, 2024.

\bibitem{cite24}
J.~Zhang, J.~Huang, S.~Jin, and S.~Lu, ``Vision-language models for vision tasks: A survey,'' \emph{IEEE Transactions on Pattern Analysis and Machine Intelligence}, vol.~46, no.~8, pp. 5625--5644, 2024.

\bibitem{cite25}
A.~Ghosh, A.~Acharya, S.~Saha, V.~Jain, and A.~Chadha, ``Exploring the frontier of vision-language models: A survey of current methodologies and future directions,'' \emph{arXiv preprint arXiv:2404.07214}, 2024.

\bibitem{cite26}
K.~Ranasinghe, X.~Li, K.~Kahatapitiya, and M.~S. Ryoo, ``Understanding long videos with multimodal language models,'' \emph{arXiv preprint arXiv:2403.16998}, 2024.

\bibitem{cite27}
J.~Min, S.~Buch, A.~Nagrani, M.~Cho, and C.~Schmid, ``Morevqa: Exploring modular reasoning models for video question answering,'' in \emph{Proceedings of the IEEE/CVF Conference on Computer Vision and Pattern Recognition}, 2024, pp. 13\,235--13\,245.

\bibitem{cite28}
Y.~Fan, X.~Ma, R.~Wu, Y.~Du, J.~Li, Z.~Gao, and Q.~Li, ``Videoagent: A memory-augmented multimodal agent for video understanding,'' in \emph{European Conference on Computer Vision}.\hskip 1em plus 0.5em minus 0.4em\relax Springer, 2024, pp. 75--92.

\bibitem{cite29}
Z.~Wang, S.~Yu, E.~Stengel-Eskin, J.~Yoon, F.~Cheng, G.~Bertasius, and M.~Bansal, ``Videotree: Adaptive tree-based video representation for llm reasoning on long videos,'' \emph{arXiv preprint arXiv:2405.19209}, 2024.

\bibitem{cite30}
B.~Wang, Y.~Li, Z.~Lv, H.~Xia, Y.~Xu, and R.~Sodhi, ``Lave: Llm-powered agent assistance and language augmentation for video editing,'' in \emph{Proceedings of the 29th International Conference on Intelligent User Interfaces}, ser. IUI '24.\hskip 1em plus 0.5em minus 0.4em\relax New York, NY, USA: Association for Computing Machinery, 2024, p. 699–714.

\bibitem{cite31}
M.~Leake and W.~Li, ``Chunkyedit: Text-first video interview editing via chunking,'' in \emph{Proceedings of the 2024 CHI Conference on Human Factors in Computing Systems}, ser. CHI '24.\hskip 1em plus 0.5em minus 0.4em\relax New York, NY, USA: Association for Computing Machinery, 2024.

\bibitem{cite35}
T.~Soucek and J.~Lokoc, ``Transnet v2: An effective deep network architecture for fast shot transition detection,'' in \emph{Proceedings of the 32nd ACM International Conference on Multimedia}, 2024, pp. 11\,218--11\,221.

\bibitem{cite34}
Y.~Li, F.~Tian, H.~Xu, and T.~Lu, ``Toward unified and quantitative cinematic shot attribute analysis,'' \emph{Electronics}, vol.~12, no.~19, p. 4174, 2023.

\bibitem{cite39}
T.~Yu, H.~Zhang, Q.~Li, Q.~Xu, Y.~Yao, D.~Chen, X.~Lu, G.~Cui, Y.~Dang, T.~He \emph{et~al.}, ``Rlaif-v: Open-source ai feedback leads to super gpt-4v trustworthiness,'' \emph{Preprint}, 2024.

\bibitem{cite36}
A.~Radford, J.~W. Kim, T.~Xu, G.~Brockman, C.~McLeavey, and I.~Sutskever, ``Robust speech recognition via large-scale weak supervision,'' in \emph{International conference on machine learning}.\hskip 1em plus 0.5em minus 0.4em\relax PMLR, 2023, pp. 28\,492--28\,518.

\bibitem{cite37}
Y.~Tsivian, \emph{Cinemetrics, part of the humanities’ cyberinfrastructure}.\hskip 1em plus 0.5em minus 0.4em\relax transcript, 2009.

\bibitem{cite38}
M.~Baxter, ``Notes on cinemetric data analysis,'' \emph{Chicago: Cinemetrics. http://www. cinemetrics. lv/dev/Cinemetrics\_Book\_Baxter. pdf}, 2014.

\bibitem{cite40}
C.~Feichtenhofer, H.~Fan, J.~Malik, and K.~He, ``Slowfast networks for video recognition,'' in \emph{Proceedings of the IEEE/CVF international conference on computer vision}, 2019, pp. 6202--6211.

\bibitem{cite41}
D.~Tran, L.~Bourdev, R.~Fergus, L.~Torresani, and M.~Paluri, ``Learning spatiotemporal features with 3d convolutional networks,'' in \emph{Proceedings of the IEEE international conference on computer vision}, 2015, pp. 4489--4497.

\bibitem{cite42}
Y.~Li, C.-Y. Wu, H.~Fan, K.~Mangalam, B.~Xiong, J.~Malik, and C.~Feichtenhofer, ``Mvitv2: Improved multiscale vision transformers for classification and detection,'' in \emph{Proceedings of the IEEE/CVF conference on computer vision and pattern recognition}, 2022, pp. 4804--4814.

\bibitem{cite43}
Z.~Liu, J.~Ning, Y.~Cao, Y.~Wei, Z.~Zhang, S.~Lin, and H.~Hu, ``Video swin transformer,'' in \emph{Proceedings of the IEEE/CVF conference on computer vision and pattern recognition}, 2022, pp. 3202--3211.

\bibitem{cite44}
H.~Liu, C.~Li, Q.~Wu, and Y.~J. Lee, ``Visual instruction tuning,'' in \emph{Advances in Neural Information Processing Systems}, A.~Oh, T.~Naumann, A.~Globerson, K.~Saenko, M.~Hardt, and S.~Levine, Eds., vol.~36.\hskip 1em plus 0.5em minus 0.4em\relax Curran Associates, Inc., 2023, pp. 34\,892--34\,916.

\bibitem{cite45}
Y.~Yao, T.~Yu, A.~Zhang, C.~Wang, J.~Cui, H.~Zhu, T.~Cai, H.~Li, W.~Zhao, Z.~He \emph{et~al.}, ``Minicpm-v: A gpt-4v level mllm on your phone,'' \emph{arXiv preprint arXiv:2408.01800}, 2024.

\bibitem{cite46}
H.~Touvron, T.~Lavril, G.~Izacard, X.~Martinet, M.-A. Lachaux, T.~Lacroix, B.~Rozi{\`e}re, N.~Goyal, E.~Hambro, F.~Azhar \emph{et~al.}, ``Llama: Open and efficient foundation language models,'' \emph{arXiv preprint arXiv:2302.13971}, 2023.

\bibitem{cite47}
E.~J. Hu, Y.~Shen, P.~Wallis, Z.~Allen-Zhu, Y.~Li, S.~Wang, L.~Wang, W.~Chen \emph{et~al.}, ``Lora: Low-rank adaptation of large language models.'' \emph{ICLR}, vol.~1, no.~2, p.~3, 2022.

\bibitem{cite50}
X.~Zheng, Y.~Li, H.~Chu, Y.~Feng, X.~Ma, J.~Luo, J.~Guo, H.~Qin, M.~Magno, and X.~Liu, ``An empirical study of qwen3 quantization,'' \emph{arXiv preprint arXiv:2505.02214}, 2025.

\bibitem{cite51}
D.~Guo, D.~Yang, H.~Zhang, J.~Song, R.~Zhang, R.~Xu, Q.~Zhu, S.~Ma, P.~Wang, X.~Bi \emph{et~al.}, ``Deepseek-r1: Incentivizing reasoning capability in llms via reinforcement learning,'' \emph{arXiv preprint arXiv:2501.12948}, 2025.

\bibitem{cite52}
F.~Caba~Heilbron, V.~Escorcia, B.~Ghanem, and J.~Carlos~Niebles, ``Activitynet: A large-scale video benchmark for human activity understanding,'' in \emph{Proceedings of the ieee conference on computer vision and pattern recognition}, 2015, pp. 961--970.

\bibitem{cite53}
Ollama, ``Ollama: Get up and running with large language models,'' \url{https://github.com/ollama/ollama}, 2025, accessed: 2025-05-16.

\bibitem{cite59}
S.~Wu, W.~Yang, and S.~Wu, ``Spatiotemporal-aware visual captioning using vision-language pre-training model,'' in \emph{ICASSP 2025 - 2025 IEEE International Conference on Acoustics, Speech and Signal Processing (ICASSP)}, 2025, pp. 1--5.

\bibitem{cite54}
A.~Arnab, M.~Dehghani, G.~Heigold, C.~Sun, M.~Lu{\v{c}}i{\'c}, and C.~Schmid, ``Vivit: A video vision transformer,'' in \emph{Proceedings of the IEEE/CVF international conference on computer vision}, 2021, pp. 6836--6846.

\bibitem{cite55}
A.~Radford, J.~W. Kim, C.~Hallacy, A.~Ramesh, G.~Goh, S.~Agarwal, G.~Sastry, A.~Askell, P.~Mishkin, J.~Clark, G.~Krueger, and I.~Sutskever, ``Learning transferable visual models from natural language supervision,'' in \emph{Proceedings of the 38th International Conference on Machine Learning}, ser. Proceedings of Machine Learning Research, M.~Meila and T.~Zhang, Eds., vol. 139.\hskip 1em plus 0.5em minus 0.4em\relax PMLR, 18--24 Jul 2021, pp. 8748--8763.

\bibitem{cite56}
T.~Chen, S.~Kornblith, M.~Norouzi, and G.~Hinton, ``A simple framework for contrastive learning of visual representations,'' in \emph{Proceedings of the 37th International Conference on Machine Learning}, ser. Proceedings of Machine Learning Research, H.~D. III and A.~Singh, Eds., vol. 119.\hskip 1em plus 0.5em minus 0.4em\relax PMLR, 13--18 Jul 2020, pp. 1597--1607.

\bibitem{cite57}
Z.~Liu, J.~Ning, Y.~Cao, Y.~Wei, Z.~Zhang, S.~Lin, and H.~Hu, ``Video swin transformer,'' in \emph{Proceedings of the IEEE/CVF conference on computer vision and pattern recognition}, 2022, pp. 3202--3211.

\bibitem{cite58}
Z.~Tong, Y.~Song, J.~Wang, and L.~Wang, ``Videomae: Masked autoencoders are data-efficient learners for self-supervised video pre-training,'' in \emph{Advances in Neural Information Processing Systems}, S.~Koyejo, S.~Mohamed, A.~Agarwal, D.~Belgrave, K.~Cho, and A.~Oh, Eds., vol.~35.\hskip 1em plus 0.5em minus 0.4em\relax Curran Associates, Inc., 2022, pp. 10\,078--10\,093.

\end{thebibliography}
\end{document}